\documentclass[11pt]{article}
\usepackage{amsfonts, amssymb, amsmath}
\usepackage{algorithm, algorithmic, amsthm}
\usepackage{graphicx}
\usepackage{color}
\usepackage{hyperref}
\usepackage[top=2.5cm, bottom=3cm, left=3cm, right=3cm]{geometry}

\usepackage{subfigure}

\usepackage{hyperref}
\usepackage{url}
\usepackage{graphicx, wrapfig}
\usepackage{amsfonts, amssymb, amsmath}
\usepackage{algorithm, algorithmic, amsthm}
\usepackage{rotating}
\usepackage{multirow}
\usepackage{paralist}
\usepackage{natbib}

\newcommand{\sss}{\ensuremath{\mathbf{s}}}  

\newcommand{\x}{\ensuremath{\mathbf{x}}}

\newcommand{\calC}{\ensuremath{\mathcal{C}}}

\newcommand{\calG}{\ensuremath{\mathcal{G}}}

\newcommand{\calO}{\ensuremath{\mathcal{O}}}

\newcommand{\calS}{\ensuremath{\mathcal{S}}}

{%
\begin{list}{#1}{
\vspace{-\topsep}
\vspace{-\partopsep}
\setlength{\itemindent}{0cm}
\setlength{\rightmargin}{0cm}
\setlength{\listparindent}{0cm}
\settowidth{\labelwidth}{#1}
\setlength{\leftmargin}{\labelwidth}
\addtolength{\leftmargin}{\labelsep}
\setlength{\itemsep}{0cm}
}%
}%
{%
\end{list}
\vspace{-\topsep}
\vspace{-\partopsep}
}

{\begin{enumerate}%
}%
{\end{enumerate}}

\theoremstyle{plain}

\newtheorem*{lemma*}{Lemma}

\newtheorem*{prop*}{Proposition}

\theoremstyle{definition}

\newtheorem*{defn*}{Definition}

\newtheorem*{exmp*}{Example}

\newtheorem*{conj*}{Conjecture}

\theoremstyle{remark}

\newtheorem*{rmk*}{Remark}

\renewcommand{\cite}[1]{\citep{#1}}

\newcommand{\oonp}{\textsf{OONP }}

\newcommand{\OONP}{\textsf{OONP }}
\newcommand{\OONPx}{\textsf{OONP}}

\newcommand{\carryon}{\textsf{Carry-on Memory }}
\newcommand{\carryonx}{\textsf{Carry-on Memory}}

\newcommand{\pppx}{\textsc{Person}}

\newcommand{\iiix}{\textsc{Item}}

\newcommand{\eeex}{\textsc{Event}}

\newcommand{\poo}{\textsc{Person}-object }
\newcommand{\poox}{\textsc{Person}-object}
\newcommand{\poos}{\textsc{Person}-objects }

\newcommand{\pclsx}{\textsc{Person}}

\newcommand{\ioo}{\textsc{Item}-object }
\newcommand{\ioox}{\textsc{Item}-object}
\newcommand{\ioos}{\textsc{Item}-objects }

\newcommand{\iclsx}{\textsc{Item}}

\newcommand{\eoo}{\textsc{Event}-object }
\newcommand{\eoos}{\textsc{Event}-objects }
\newcommand{\eoox}{\textsc{Event}-object}

\newcommand{\eclsx}{\textsc{Event}}

\newcommand{\loox}{\textsc{Location}-object}

\newcommand{\tina}{\texttt{Tina} }
\newcommand{\tinax}{\texttt{Tina}}
\newcommand{\apple}{\texttt{apple} }
\newcommand{\applex}{\texttt{apple}}
\newcommand{\kitchen}{\texttt{kitchen} }

\newcommand{\garden}{\texttt{garden} }

\newcommand{\timee}{\texttt{Time} }

\newcommand{\location}{\texttt{Location} }
\newcommand{\typeex}{\texttt{Type}}
\newcommand{\typee}{\texttt{Type} }

\newcommand{\modelx}{\texttt{Model}}
\newcommand{\colorrx}{\texttt{Color}}

\newcommand{\statusx}{\texttt{Status}}

\newcommand{\desc}{\texttt{Description} }

\newcommand{\returned}{\texttt{Returned}}

\newcommand{\age}{\texttt{Age} }
\newcommand{\name}{\texttt{Name} }
\newcommand{\gender}{\texttt{Gender} }

\newcommand{\quantity}{\texttt{Quantity} }
\newcommand{\valuee}{\texttt{Value} }
\newcommand{\valuex}{\texttt{Value}}

\newcommand{\namex}{\texttt{Name}}

\newcommand{\theftx}{\texttt{theft}}

\newcommand{\restix}{\texttt{restitutionx}}

\newcommand{\dispx}{\texttt{disposal}}

\newcommand{\buyerx}{\texttt{buyer}}
\newcommand{\companionx}{\texttt{companion}}
\newcommand{\principalx}{\texttt{principal}}

\newcommand{\suspect}{\texttt{suspect} }

\newcommand{\carryx}{\texttt{carry}}

\newcommand{\carrybyx}{\texttt{iscarriedby}}

\newcommand{\locatetightx}{\texttt{islocatedat}}
\newcommand{\locatex}{\texttt{is-located-at}}

\newcommand{\suspectx}{\texttt{suspect}}
\newcommand{\victimx}{\texttt{victim}}

\newcommand{\inline}{\textsf{Inline Memory }}
\newcommand{\inlinex}{\textsf{Inline Memory}}
\newcommand{\minline}{$\textsf{M}_\text{inl}$ }

\newcommand{\minlinet}{$\textsf{M}_\text{inl}^t$ }
\newcommand{\minlinetx}{$\textsf{M}_\text{inl}^t$}

\newcommand{\om}{\textsf{Object Memory }}
\newcommand{\omx}{\textsf{Object Memory}}
\newcommand{\mom}{$\textsf{M}_\text{obj}$ }
\newcommand{\momx}{$\textsf{M}_\text{obj}$}
\newcommand{\momt}{$\textsf{M}_\text{obj}^t$ }
\newcommand{\momtx}{$\textsf{M}_\text{obj}^t$}

\newcommand{\mat}{\textsf{Matrix Memory }}
\newcommand{\mmat}{$\textsf{M}_\text{mat}$ }
\newcommand{\mmatt}{$\textsf{M}_\text{mat}^t$ }
\newcommand{\mmattx}{$\textsf{M}_\text{mat}^t$}
\newcommand{\mmatx}{$\textsf{M}_\text{mat}$}

\newcommand{\act}{\textsf{Action History }}
\newcommand{\mactx}{$\textsf{M}_\text{act}$}

\newcommand{\macttx}{$\textsf{M}_\text{act}^t$ }

\newcommand{\matx}{\textsf{Matrix Memory}}
\newcommand{\actx}{\textsf{Action History}}
\newcommand{\nnc}{\textsf{Neural Net Controller }}

\newcommand{\nncsx}{\textsf{NNC}}
\newcommand{\nncx}{\textsf{Neural Net Controller}}
\newcommand{\pn}{\textsf{Policy-net }}
\newcommand{\pnx}{\textsf{Policy-net}}

\newcommand{\sr}{\textsf{Symbolic Reasoner }}
\newcommand{\srx}{\textsf{Symbolic Reasoner}}
\newcommand{\sm}{\textsf{Symbolic Matching} }
\newcommand{\smx}{\textsf{Symbolic Matching}}

\newcommand{\sax}{\textsf{Symbolic Analyzer}}

\newcommand{\Reader}{\textsf{Reader }}
\newcommand{\Readerx}{\textsf{Reader}}

\newcommand{\newassign}{\texttt{New-Assign} }
\newcommand{\neww}{\texttt{New} }

\newcommand{\newassignx}{\texttt{New-Assign}}

\newcommand{\assign}{\texttt{Assign} }
\newcommand{\assignx}{\texttt{Assign}}

\newcommand{\updatewhat}{\texttt{Update.X} }
\newcommand{\updatetowhat}{\texttt{Update2what} }
\newcommand{\updatewhatx}{\texttt{Update.X}}
\newcommand{\updatetowhatx}{\texttt{Update2what}}

\newcommand{\none}{\texttt{none} }

\renewcommand{\thefootnote}{\fnsymbol{footnote}}

\title{ Object-oriented Neural Programming (OONP) \\ for Document Understanding}

\author{\sf Zhengdong Lu$^{1}$,
	Xianggen Liu$^{2,3,4,*}$,
	Haotian Cui$^{2,3,4,*}$, Yukun Yan$^{2,3,4,*}$  
	Daqi Zheng$^1$\\
	\tt {luz@deeplycurious.ai,} \\ \tt{\{liuxg16,cht15,yanyk13\}@mails.tsinghua.edu.cn, 	da@deeplycurious.ai}\\
	$^1$ DeeplyCurious.ai\\
	$^2$ Department of Biomedical Engineering, School of Medicine, Tsinghua University\\ 
	$^3$ Beijing Innovation Center for Future Chip, Tsinghua University\\
	$^4$ Laboratory for Brain and Intelligence, Tsinghua University
}

\date{}
\begin{document}
\maketitle

\begin{abstract}
	We propose \textsf{\small Object-oriented Neural Programming} (\OONPx),  a framework for semantically parsing documents in specific domains.  Basically, \OONP reads a document and parses it into a predesigned object-oriented data structure  that reflects the domain-specific semantics of the document.  An \oonp parser models semantic parsing as a decision process: a neural net-based Reader sequentially goes through the document, and builds and updates an intermediate ontology   during the process to summarize its partial understanding of the text. \oonp supports a big variety of forms  (both symbolic and differentiable) for representing the state and the document, and 
	a rich family of operations to compose the representation.  An \oonp parser can be trained with supervision of different forms and strength,  including supervised learning (SL) , reinforcement learning (RL) and hybrid of the two.  Our experiments on both synthetic and real-world document parsing tasks have shown that \OONP can learn to handle fairly complicated ontology with training data  of modest sizes.\let\thefootnote\relax\footnotetext{* The work was done when these authors worked as interns at DeeplyCurious.ai.}
\end{abstract}

\section{Introduction}
Mapping a document into a structured ``machine readable" form is a canonical and probably the most effective way for document understanding.  There are quite some recent efforts on designing neural net-based learning machines for this purpose, which can be roughly categorized into two groups:  1) sequence-to-sequence model with the neural net as the the black box~\cite{Dong16-seq2seq-from-lili, Liang17-seq2seq-from-lili}, and 2)  neural net as a component in a pre-designed statistical model~\cite{Zeng14-relationClassification}.  We however argue that both approaches have their own serious problems and cannot be used on document with relatively complicated structures.   Towards solving this problem,  we proposed \textsf{Object-oriented Neural Programming} (\OONP),  a framework for semantically parsing in-domain documents.  \OONP  is neural net-based, but it also has sophisticated architecture and mechanism designed for taking and outputting discrete structures, hence nicely combining symbolism (for interpretability and formal reasoning) and connectionism (for flexibility and learnability).  This ability, as we argue in this paper, is critical to document understanding. 

\begin{figure}[h!]
\centering
\includegraphics[width=0.98\textwidth]{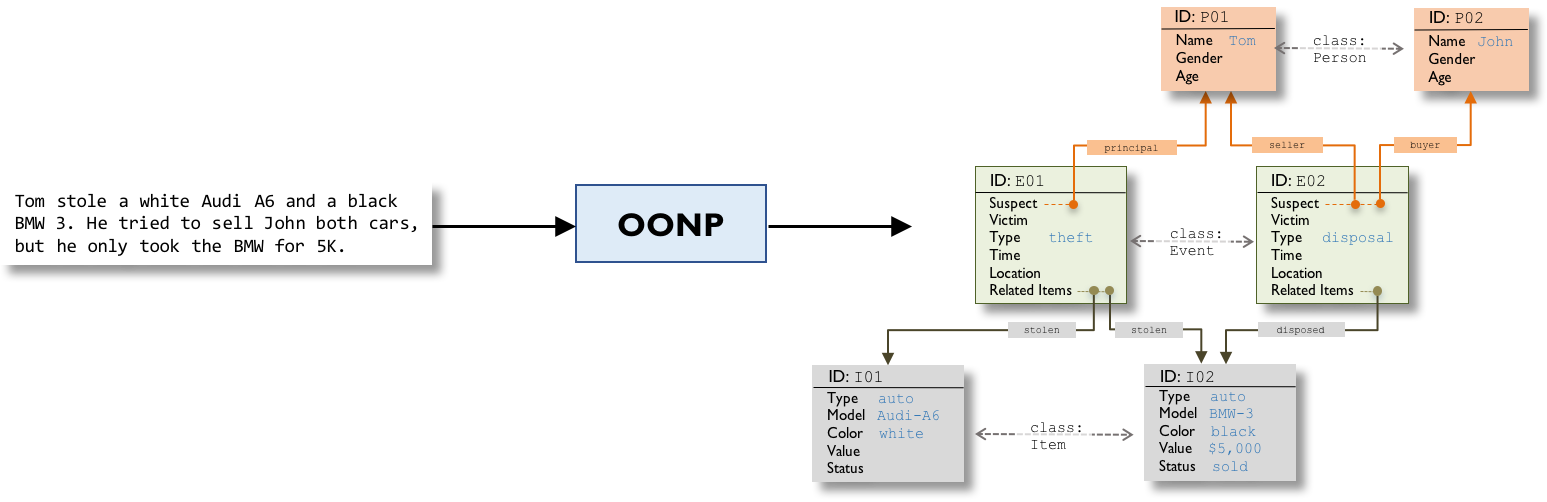}
\caption{ Illustration of \OONP on a parsing task.}
\label{f:OONPatWork}
\end{figure}

\oonp seeks to map a document to a graph structure with each node being an object,  as illustrated in Figure \ref{f:OONPatWork}.  We borrow the name from Object-oriented Programming~\cite{Mitchell03-OOP} to emphasize the central position of ``objects" in our parsing model: indeed, the representation of objects in \oonp allows neural and symbolic reasoning over complex structures and hence it make it possible to represent much richer semantics.  Similar to Object-oriented Programming,   \oonp has the concept of ``class" and ``objects" with the following analogousness: 1) each class defines the types and organization of information it contains,  and we can define inheritance for class with different abstract levels as needed;  2) each object is an instance of a certain class, encapsulating  a number of properties and operations; 3) objects can be connected with relations (called \emph{links}) with pre-determined type.   Based on objects, we can define ontology and operations that reflect the intrinsic structure of the parsing task.  

For parsing, \OONP reads a document and parses it into this object-oriented data structure through a series of discrete actions along reading the document sequentially.  \oonp supports a rich family of operations for composing the ontology,  and flexible hybrid forms for knowledge representation.  An \oonp parser can be trained with supervised learning (SL) , reinforcement learning (RL) and hybrid of the two.  Our experiments on one synthetic dataset and two real-world datasets have shown the efficacy of \oonp on document understanding tasks with a variety of characteristics.

\section{Related Works} 
\subsection{Semantic Parsing}
Semantic parsing is concerned with translating language utterances into executable logical forms and plays a key role in building conversational interfaces~\cite{Berant14SemanticParsing}. Different from common tasks of semantic parsings, such as parsing the sentence to dependency structure~\cite{buys2017robust} and executable commands~\cite{herzig2017neural}, \OONP parses documents into a predesigned object-oriented data structure which is easily readable for both human and machine. It is related to semantic web~\cite{berners2001semantic} as well as frame semantics~\cite{Fillmore1982frame}  in the way semantics is represented, so in a sense, \OONP can be viewed as a neural-symbolic implementation of semantic parsing with similar semantic representation.  
\subsection{State Tracking}
\OONP is inspired by~\citet{Daume09-searn} on modeling parsing as a decision process, and the  work on  state-tracking models in dialogue system~\cite{Henderson14-state-tracking} for the mixture of symbolic and probabilistic representations of dialogue state. 
For modeling a document with entities,  \citet{yang2016reference} use coreference links to recover entity clusters, though they only model entity mentions as containing a single
word. However, entities whose names consist of multiple words are not considered.  Entity Networks~\cite{Henaff16-entityNetwork} and \textsf{\small EntityNLM}~\cite{ji2017entitynlm} have addressed above problem and are the pioneers to model on tracking entities, but they have  not considered the properties of the entities. In fact, explicitly modeling the entities both with their properties and contents is important to understand a document, especially a complex document. For example, if there are two persons named `Avery', it is vital to know their genders or last names to avoid confusion. Therefore, we propose \OONP to sketch objects and their relationships by building a structured graph for document parsing.


\section{Overview of \OONP}
\label{s:overview}
\begin{figure}[h!]
\centering
\includegraphics[width=0.92\textwidth]{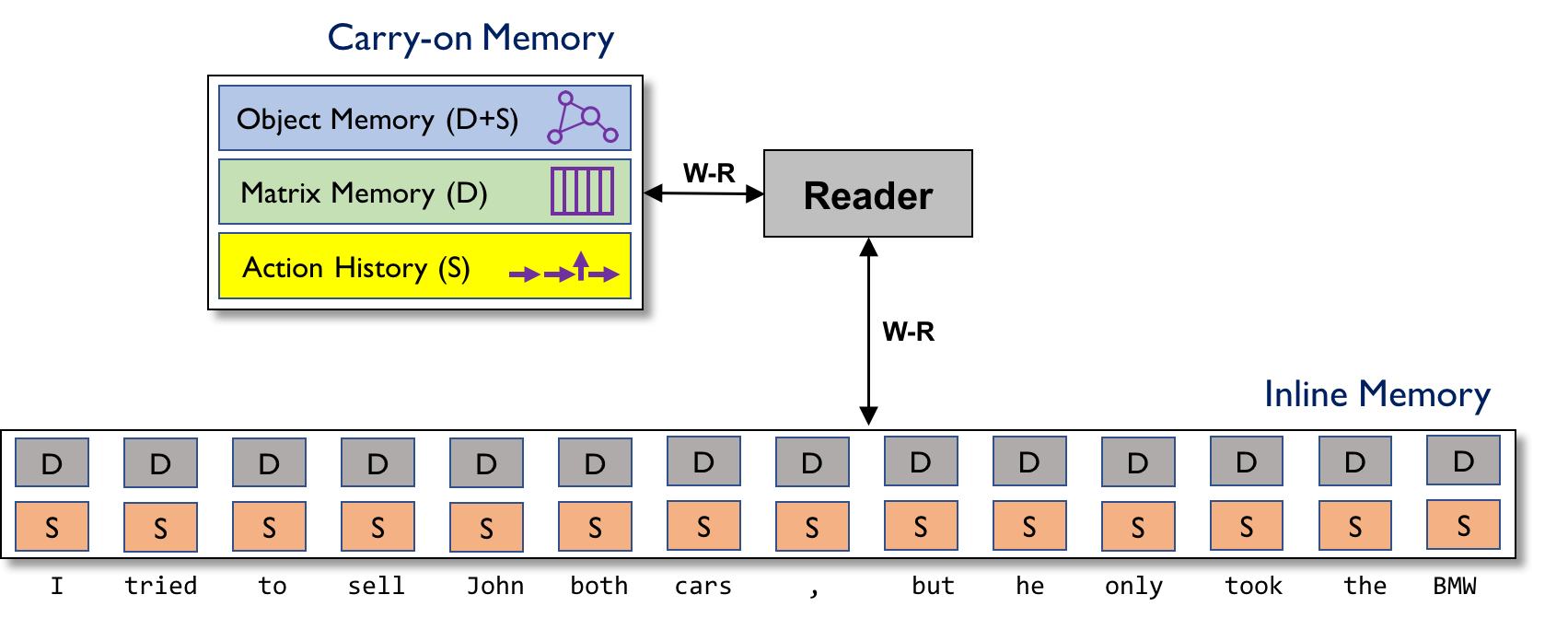}
\caption{ The overall digram of \OONPx, where S stands for symbolic representation, D stands for distributed representation, and S+D stands for a hybrid representation with both symbolic and distributed parts.}
\label{f:OONPdiagram}
\end{figure}

An \OONP parser (as illustrated through the diagram in Figure \ref{f:OONPdiagram}) consists of a \Reader equipped with read/write heads,  \inline that represents the document,  and \carryon that summarizes the current understanding of the document at each time step.  For each document to parse, \OONP first preprocesses it and puts it into the \inline, and then \Reader controls the read-heads to sequentially go through the \inline (for possibly multiple times, see Section \ref{s:e3} for an example) and at the same time update the \carryonx.    

The major components of \OONP are described in the following: \vspace{-5pt}
\begin{itemize}

\item {\bf Memory:}  we have two types of memory,  \carryon and \inlinex.
\carryon is designed to save the state\footnote{It is not entirely accurate, since the \inline can be modified during the reading process it also records some of the state information.} in the decision process and summarize current understanding of the document based on the text that has been `read".  \carryon has three compartments: 
\begin{itemize}
\item \omx :  denoted as \momx, the object-based ontology constructed during the parsing process, see Section \ref{s:om} for details;
\item \matx :   denoted as \mmatx,  a matrix-type memory with fixed size, for differentiable read/write by the controlling neural net~\cite{Graves14-NTM}. In the simplest case, it could be just a vector as the hidden state of conventional Recurrent Neural Netwokr (RNN);
\item \actx : denoted as \mactx, saving the entire history of actions made during the parsing process.
\end{itemize}
Intuitively,  \mom stores the extracted knowledge with defined structure and strong evidence,  while  \mmat keeps the knowledge that is fuzzy, uncertain or incomplete, waiting for future information to confirm, complete and clarify.  \inline, denoted \minline ,  is designed to save location-specific information about the document. In a sense, the information in \inline is low level and unstructured, waiting for \Reader to fuse and integrate for more structured representation.

\item {\bf Reader:}  \Reader is the control center of \OONPx,  coordinating and managing all the operations of \OONPx.  More specifically,  it takes the input  of different forms (reading), processes it (thinking),  and updates the memory (writing).   As shown in Figure \ref{f:Reader1},  \Reader contains \nnc (\nncsx) and multiple symbolic processors, and \nnc also has \pn as its sub-component.  Similar to the controller in Neural Turing Machine~\cite{Graves14-NTM}, \nnc is equipped with multiple read-heads and write-heads for differentiable read/write over \mat and (the distributed part of) \inlinex, with possibly a variety of addressing strategies~\cite{Graves14-NTM}.   \pn however issues discrete outputs (i.e., \emph{actions}),  which gradually builds and updates the \om in time (see Section \ref{s:om} for more details). The actions could also updates the symbolic part of \inline if needed.  The symbolic processors are designed to handle information in symbolic form from \omx, \inlinex, \actx, and  \pnx,  while that from \inline and \act  is eventually generated by \pnx.
\begin{figure}[h!]
\centering
\includegraphics[width=0.4\textwidth]{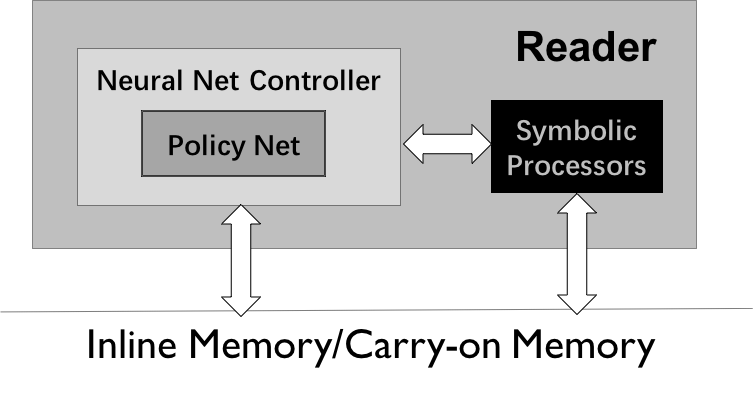}
\caption{ The overall digram of \OONP}
\label{f:Reader1}
\end{figure}

\end{itemize}

We can show how the major components of \OONP collaborate to make it work through the following sketchy example.  In reading the following text\\ \vspace{-20pt}
\begin{figure*}[h!]
\centering
\includegraphics[width=0.9\textwidth]{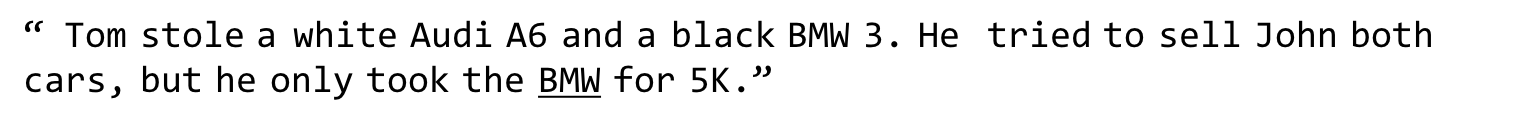} \vspace{-20pt}
\end{figure*}  
\\  \OONP has reached the underlined word ``BMW" in \inlinex.  At this moment,  \OONP has two objects (\texttt{I01} and \texttt{I02})  for Audi-06 and BMW respectively in \omx.  \Reader determines that the information it is currently holding  is about \texttt{I02} (after comparing it with both objects) and updates its status property to  \texttt{sold},  along with other update on both \mat and \actx.

\paragraph{\OONP in a nutshell:} The key properties of \OONP can be summarized as follows  \vspace{-4pt}
\begin{enumerate}
\item \OONP models parsing as a  decision process:  as the ``reading and comprehension" agent goes through the text it gradually forms the ontology as the representation of the text through its action;  \vspace{-4pt}
\item \OONP uses a symbolic memory with graph structure as part of the state of the parsing process. This memory will be created and updated through the sequential actions of the decision process,  and will be used as the semantic representation of the text at the end;  \vspace{-4pt}
\item \OONP can blend supervised learning (SL) and reinforcement learning (RL) in tuning its parameters to suit the supervision signal in different forms and strength;  \vspace{-4pt}
\item \OONP allows different ways to add symbolic knowledge into the raw representation of the text (\inlinex) and its policy net in forming the final structured representation of the text.  \vspace{-4pt}
\end{enumerate}


\section{\OONPx : Components} \label{s:components}
In this section we will discuss the major components in \OONPx, namely \om, \inline  and \Readerx.   We omit the discussion on \mat and \act since they are straightforward given the description in Section~\ref{s:overview}.

\subsection{\om} \label{s:om}
\begin{figure}[h!]
\centering
\includegraphics[width=0.7\textwidth]{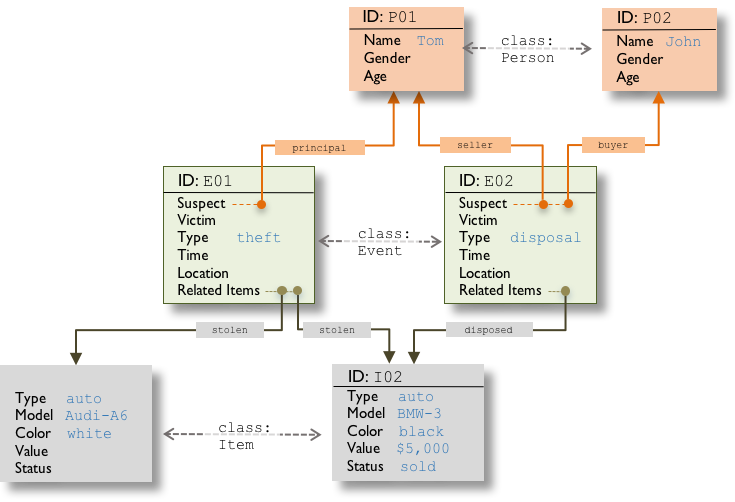}
\caption{ An example of the objects from three classes.}
\label{f:OntExample}
\end{figure}

\om stores an object-oriented representation of document, as illustrated in Figure~\ref{f:OntExample}.   Each object is an instance of a particular class\footnote{In this paper, we limit ourselves to a flat structure of classes, but it is possible and even beneficial to have a hierarchy of classes. In other words, we can have classes with different levels of abstractness, and  allow an object to go from abstract class to its child class during the parsing process, with more and more information is obtained. }, which specifies the internal structure of the object, including internal properties, operations, and how this object can be connected with others. The internal properties can be of different types, for example string or category, which usually correspond to different actions in composing them:  the string-type property is usually ``copied" from the original text in \inlinex, while the category properties usually needs to be rendered by a classifier. The links are by nature bi-directional, meaning that it can be added from both ends (e.g., in the experiment in Section \ref{s:babi}), but for modeling convenience, we might choose to let it to be one directional (e.g., in the experiments in Section \ref{s:e2} and \ref{s:e3}).  In Figure~\ref{f:OntExample}, there are six ``linked" objects of three classes (namely, \pclsx, \eclsx, and \iclsx) .  Taking \ioo \texttt{I02} for example,  it has five internal properties (\typeex, \modelx, \colorrx, \valuex, \statusx ), and is linked with two \eoox s through \texttt{stolen} and \texttt{disposed} link respectively.

In addition to the symbolic part,  each object had also its own distributed presentation (named \emph{object-embedding}), which serves as its interface with other distributed representations in \Reader (e.g., those from the \mat or the distributed part of \inlinex). For description simplicity, we will refer to the symbolic part of this hybrid representation of objects as  \emph{ontology}, with some slight abuse of this word. Object-embedding serves as a dual representation to the symbolic part of a  object, recording all the relevant information associated with it but not represented in the ontology, e.g., the context of text when the object is created.   

The representations  in \omx,   including the ontology and object embeddings, will be updated in time by the operations defined for the corresponding classes. Usually, the actions are the driving force in those operations, which not only  initiate and grow the ontology, but also coordinate other differentiable operations.  For example, object-embedding associated with a certain object  changes with any non-trivial action concerning this object, e.g., any update on the internal properties or the external links, or even a mention (corresponding to an \assign action described in Section~\ref{s:actions}) without any update.  

According to the way the ontology evolves with time,   the parsing task can be roughly classified into two categories
\begin{itemize}
\item {\bf Stationary: }  there is a final ground truth that does not change with time.  So with any partial history of the text,  the corresponding ontology is always part of the final one, while the missing part is due to the lack of information.  See task in Section~\ref{s:e2} and \ref{s:e3} for example.
\item {\bf Dynamical: } the truth changes with time,  so the ontology corresponding to partial history of text may be different from that of the final state. See task in Section~\ref{s:babi} for example.
\end{itemize}
It is important to notice that this categorization depends not only on the text but also  heavily on the definition of ontology.   Taking the text in Figure~\ref{f:OONPatWork} for example:  if we define ownership relation between a \poo and \ioox, the ontology becomes dynamical, since ownership of the BMW changed from \texttt{Tom} to \texttt{John}.

\subsection{\inline}
\inline stores the relatively raw representation of the document that follows the temporal structure of the text,  as illustrated through Figure~\ref{f:OONPdiagram}.  Basically,  \inline is an array of memory cells,  each corresponding to a pre-defined language unit (e.g., word) in the same order as they are in the original text.  Each cell can have distributed part and symbolic part, designed to save 1) the result of preprocessing of text from different models,  and 2) certain output from \Readerx,  for example from previous reading rounds.  Following are a few examples for preprocessing 
\begin{itemize}
\item  {\bf Word embedding:}  context-independent vectorial representation of words
\item {\bf Hidden states of NNs}: we can put the context in local representation of words through gated RNN like  LSTM~\cite{Greff15-LSTM} or GRU~\cite{Cho14-GRU},  or particular design of convolutional neural nets (CNN)~\cite{Yu15-dilated-conv}.  
\item {\bf Symbolic preprocessing:}  this refer to a big family of methods that yield symbolic result, including various sequential labeling models and rule-based methods. As the result we may have tag on words, extracted sub-sequences, or even relations on two pieces.
\end{itemize}
During the parsing process,   \Reader can write to \inline with its discrete or continuous outputs, a process we named \emph{``notes-taking"}.  When the output is continuous,  the notes-taking process is similar to the interactive attention in machine translation~\cite{Meng16-interactive}, which is from a  NTM-style write-head~\cite{Graves14-NTM} on \nncx.  When the output is discrete,   the notes-taking is essentially an action issued  by \pnx.

\begin{figure}[h!]\
\centering
\includegraphics[width=0.45\textwidth]{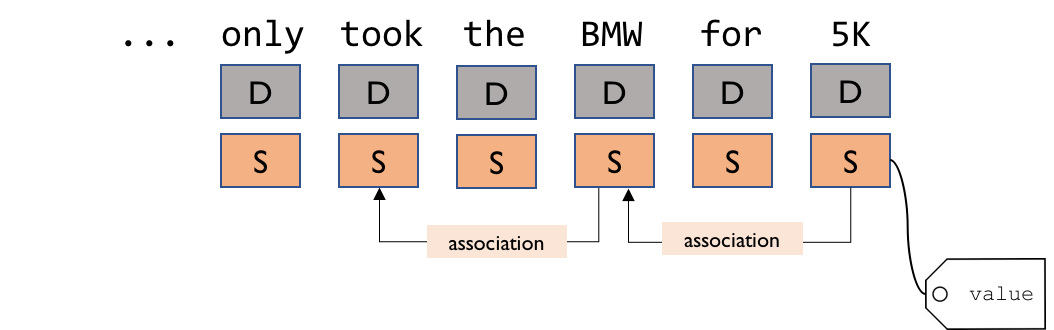}\hspace{25pt} \includegraphics[width=0.45\textwidth]{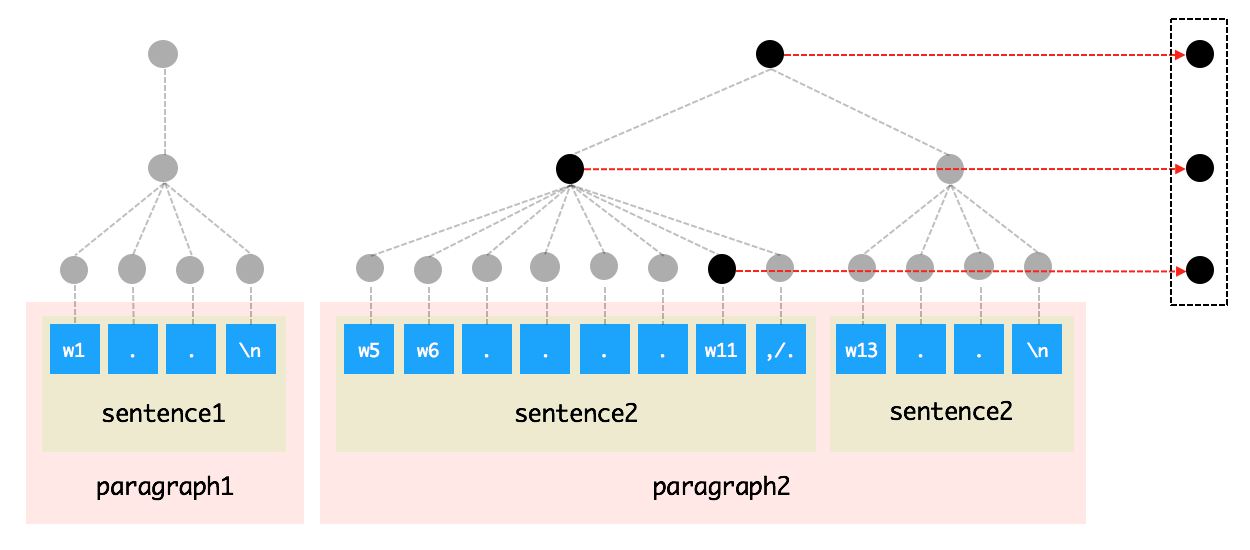}

\caption{ Left panel: \inline with symbolic knowledge;  Right panel:  one choice of nonlinear representation of the distributed part onf \inline used in~\cite
{yanyukun17}.}
\label{f:inlinerich}
\end{figure}

\inline provides a way to represent locally encoded  ``low level"  knowledge of the text, which will be read, evaluated and combined with the global semantic representation in \carryon by \Readerx.  One particular advantage of this setting is that it allows us to incorporate the local decisions of some other models, including ``higher order" ones like local relations across  two language units, as illustrated in the left panel of Figure \ref{f:inlinerich}. We can also have a rather ``nonlinear"  representation of the document in \inlinex.  As a particular example~\cite{yanyukun17},  at each location,  we can have the representation of the current word,  the representation of the rest of the sentence, and the representation of the rest of the current paragraph, which enables \Reader  to see information of  history and future at different scales, as illustrated in the right panel of Figure \ref{f:inlinerich}.

\subsection{Reader}
\begin{figure}[h!]
\centering
\includegraphics[width=0.75\textwidth]{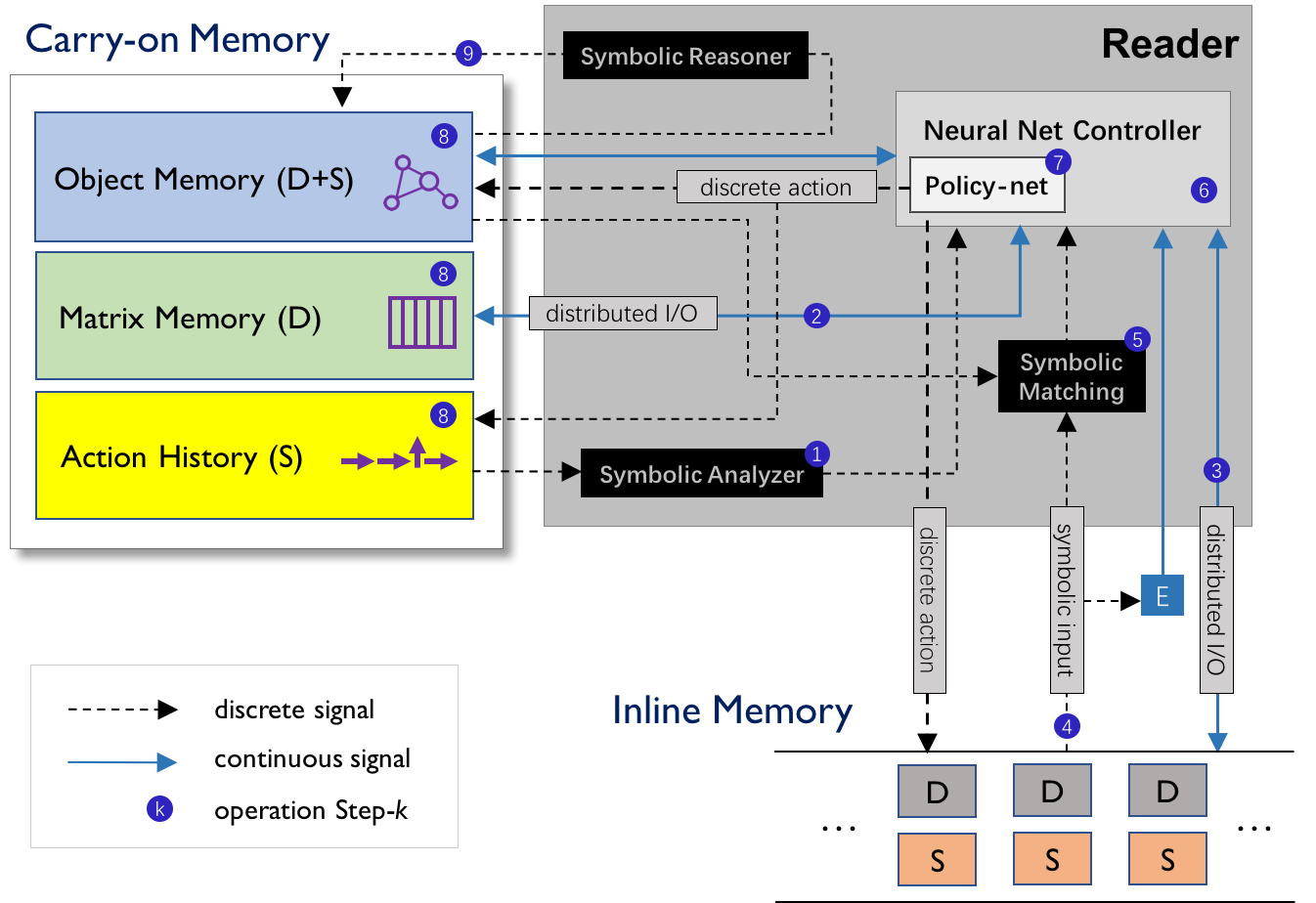}
\caption{ A particular implementation of Reader in a closer look, which reveals some details about the  entanglement of neural and symbolic components. Dashed lines stand for continuous signal and the solid lines for discrete signal}
\label{f:Reader2}
\end{figure}

\Reader is the control center of \OONP, which manages all the (continuous and discrete) operations in the \OONP parsing process.  \Reader has three symbolic processors (namely, \smx, \srx, \sax) and a \nnc  (with \pn as the sub-component). All the components in \Reader  are coupled through intensive exchange of information as shown in Figure \ref{f:Reader2}.   Below is a snapshot of the information processing at time $t$ in \Reader
{ \small
\begin{itemize}
\item \textbf{STEP-1:} let the processor \textsf{Symbolic Analyzer} to check the \act (\macttx) to  construct some symbolic features for the trajectory of actions; \vspace{-4pt}
\item \textbf{STEP-2:} access \mat (\mmattx) to get an vectorial representation for time $t$,  denoted as  $\sss_t$ ; \vspace{-4pt}
\item \textbf{STEP-3:} access \inline (\minlinetx) to get the symbolic representation $\x_t^{(s)}$ (through location-based addressing) and distributed representation $\x^{(d)}_t$ (through location-based addressing and/or content-based addressing); \vspace{-4pt}
\item \textbf{STEP-4:}  feed $\x_t^{(d)}$ and the embedding of $\x_t^{(s)}$  to \nnc to fuse with $\sss_t$; \vspace{-4pt}
\item \textbf{STEP-5:}  get the  candidate objects (some may have been eliminated by $\x_t^{(s)}$) and let them meet $\x_t^{(d)}$ through the processor  \textsf{Symbolic Matching}  for the matching of them on symbolic aspect; \vspace{-4pt}
\item \textbf{STEP-6:}  get the  candidate objects (some may have been eliminated by $\x_t^{(s)}$) and let them meet the result of STEP-4  in \nnc ; \vspace{-4pt}
\item \textbf{STEP-7:} \pn combines the result of STEP-6 and STEP-5,  to issue actions; \vspace{-4pt}
\item \textbf{STEP-8:} update \momtx, \mmatt and \minlinet with actions on both symbolic and distributed representations;  \vspace{-4pt}
\item \textbf{STEP-9:} put \momt through the processor \textsf{Symbolic Reasoner} for some high-level reasoning and logic consistency. \vspace{-4pt}
\end{itemize}
}
Note that we consider only single action for simplicity, while in practice it is common to have multiple actions at one time step, which requires a slightly more complicated design of the policy as well as the processing pipeline.

\section{\OONPx: Actions} \label{s:actions}
The actions issued by \pn can be generally categorized as the following 
\begin{itemize}
\item \newassign : determine whether to create an new object (a ``\emph{New}" operation)   for the information at hand or assign it to a certain existed object
\item \updatewhat : determine which internal property or external link of the selected object to update;
\item \updatetowhat :  determine the content of the updating, which could be about string, category or links.
\end{itemize}
The typical order of actions is \newassignx $\;\rightarrow\;$\updatewhatx$\;\rightarrow\;$\updatetowhatx, but it is very common to have \newassign action followed by nothing, when, for example, an object is mentioned but no substantial information is provided,

\subsection{\newassign}
\begin{figure}[h!]
\centering
\includegraphics[width=0.6\textwidth]{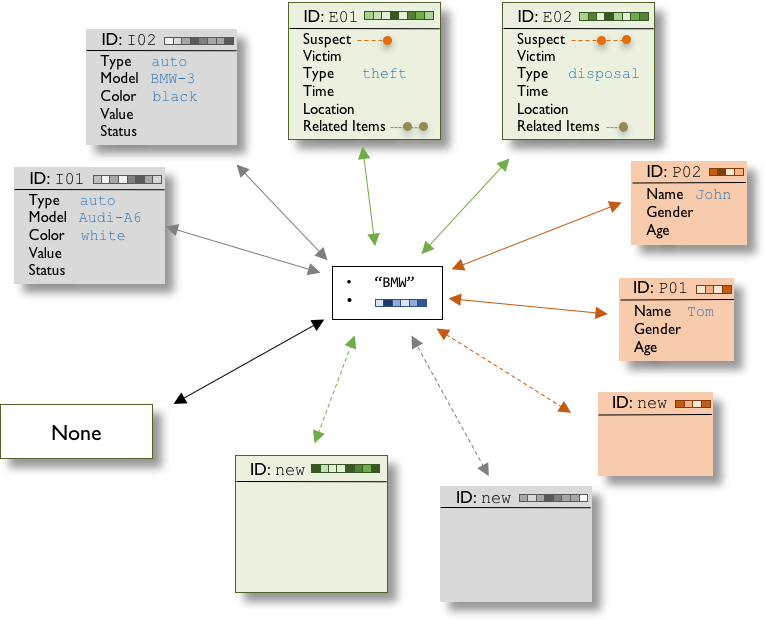}
\caption{ A pictorial illustration of what the \Reader sees in determining whether to New an object and the relevant object when the read-head on \inline reaches the last word in the sentence in Figure~\ref{f:OONPdiagram}.  The color of the arrow line stands for different matching functions for object classes, where the dashed lines is for the new object.}
\label{f:compare}
\end{figure}

With any information at hand (denoted as $\calS_t$) at time $t$,  the choices of  \newassign typically include the following three categories of actions: 1)  creating (New) an object of a certain type,  2) assigning $\calS_t$ to an existed object,  and 3) doing nothing for $\calS_t$ and moving on. For \pnx, the stochastic policy is to determine the following probabilities:
\begin{eqnarray*}
  \textsf{prob}(c,\textsf{new}| \calS_t) ,\;  && c = 1,2,\cdots, |\calC| \\
  \textsf{prob}(c, k| \calS_t) ,\;  && \text{for } \calO^{c,k}_t  \in \textsf{M}^{t}_\text{obj}\\
    \textsf{prob}(\texttt{none}| \calS_t) \;  &&  
\end{eqnarray*}
where $|\calC|$ stands for the number of classes,  $\calO^{c,k}_t$ stands for the $k^\text{th}$ object of class $c$ at time $t$.  Determining whether to new objects always relies on the following two signals
\begin{enumerate}
\item The information at hand cannot be contained by any existed objects;
\item Linguistic hints that suggests whether a new object is introduced.
\end{enumerate}
Based on those intuitions,  we takes a score-based approach to determine the above-mentioned probability.  More specifically,  for a given $\calS_t$,  \Reader forms a ``temporary" object with its own structure (denoted $\hat{\calO}_t$), including symbolic and distributed sections. In addition, we also have a virtual object for the New action for each class $c$, denoted $\calO^{c,\textsf{new}}_t$,  which is typically a time-dependent vector formed by \Reader based on information  in \mmattx .  For a given  $\hat{\calO}_t$, we can then define the following $|\calC| + |\textsf{M}^{t}_\text{obj}| + 1$ types of score functions, namely
\begin{eqnarray*}
\text{New an object of class $c$:}&&\textsf{score}^{(c)}_\textsf{new}(\calO^{c,\textsf{new}}_t,  \hat{\calO}_t; \theta_\textsf{new}^{(c)}), \;\;\;c = 1, 2, \cdots, |\calC|\;\;\;\; \\
\text{Assign to existed objects:}&& \textsf{score}^{(c)}_\textsf{assign}(\calO^{c,k}_t,  \hat{\calO}_t; \theta^{(c)}_\textsf{assign}), \; \;\; \text{for } \calO^{c,k}_t  \in \textsf{M}^{t}_\text{obj} \\
\text{Do nothing:} && \textsf{score}_\textsf{none}(\hat{\calO}_t; \theta_\textsf{none}).
\end{eqnarray*}
to measure the level of matching between the information at hand and existed objects, as well as the likeliness for creating an object or doing nothing. This process is pictorially illustrated in Figure~\ref{f:compare}.
%
%
%
%
%
We therefore can define the following probability for the stochastic policy 
\begin{eqnarray*}
\small
 \textsf{prob}(c,\textsf{new}| \calS_t)  & =  & 
  \frac{ 
  e^{\textsf{score}^{(c)}_\textsf{new}(\calO^{c,\textsf{new}}_t,  \hat{\calO}_t; \theta_\textsf{new}^{(c)})}
  }
  {Z(t)} \\  
  \textsf{prob}(c, k| \calS_t) & = &
\frac{ 
e^{\textsf{score}^{(c)}_\textsf{assign}(\calO^{c,k}_t,  \hat{\calO}_t; \theta^{(c)}_\textsf{assign})}
}
{Z(t)} \\  
\textsf{prob}(\texttt{none}| \calS_t) & = &
\frac{ 
e^{\textsf{score}_\textsf{none}(\hat{\calO}_t; \theta_\textsf{none})}
}
{
Z(t)
}  
\end{eqnarray*}
where $Z(t) =  \sum_{c' \in \calC} e^{\textsf{score}^{(c')}_\textsf{new}(\calO^{c',\textsf{new}}_t,  \hat{\calO}_t; \theta_\textsf{new}^{(c')})}+  \sum_{(c'',k')\in \textsf{idx}(\textsf{M}^{t}_\text{obj})}e^{\textsf{score}^{(c'')}_\textsf{assign}(\calO^{c'',k}_t,  \hat{\calO}_t; \theta^{(c'')}_\textsf{assign})}  + e^{\textsf{score}_\textsf{none}(\hat{\calO}_t; \theta_\textsf{none})}
$ is the normalizing factor.

Many actions are essentially trivial on the symbolic part, for example, when \pn chooses \none in \newassignx, or assigns the information at hand to an existed object but choose to update nothing in \updatewhatx, but this action will affect the distributed operations in \Readerx. This distributed operation will affect the representation in \mat or the object-embedding in \omx.

\subsection{Updating objects: \updatewhat and \updatetowhat }
In \updatewhat step,  \pn needs to choose the property or external link (or none) to update for the selected object determined by  \newassign step. If \updatewhat chooses to update an external link,  \pn needs to further determine which object it links to.   After that \updatetowhat updates the chosen property or links.  In task with static  ontology, most internal properties  and links will be ``locked" after they are updated for the first time,  with some exception on a few semi-structured property (e.g., the \texttt{Description} property in the experiment in Section \ref{s:e2}).  For dynamical ontology, on contrary, many important properties and  links are always subject to changes. A link can often be determined from both ends, e.g.,  the link that states the fact that ``{\color{blue}  \tina (a \poo)  carries \apple (an \ioo)}"  can be either specified from  from  \tinax (through adding the link ``\carryx" to \applex) or from \apple (through adding the link ``\carrybyx" to \tina), as in the experiment in Section \ref{s:babi}.  In practice, it is often more convenient to make it asymmetrical  to reduce the size of action space. 

 In practice, for a particular type of ontology, both  \updatewhat and \updatetowhat  can often be greatly simplified: for example,  
\begin{itemize}
\item when the selected object (in \newassign step) has only one property ``unlocked",  the \updatewhat step will be trivial;
\item in $\calS_t$, there is often information from \inline that tells us the basic type of the current information, which can often automatically decide the property or link.
\end{itemize}

\subsection{An example}
\begin{figure}[h!]
\centering
\includegraphics[width=0.99\textwidth]{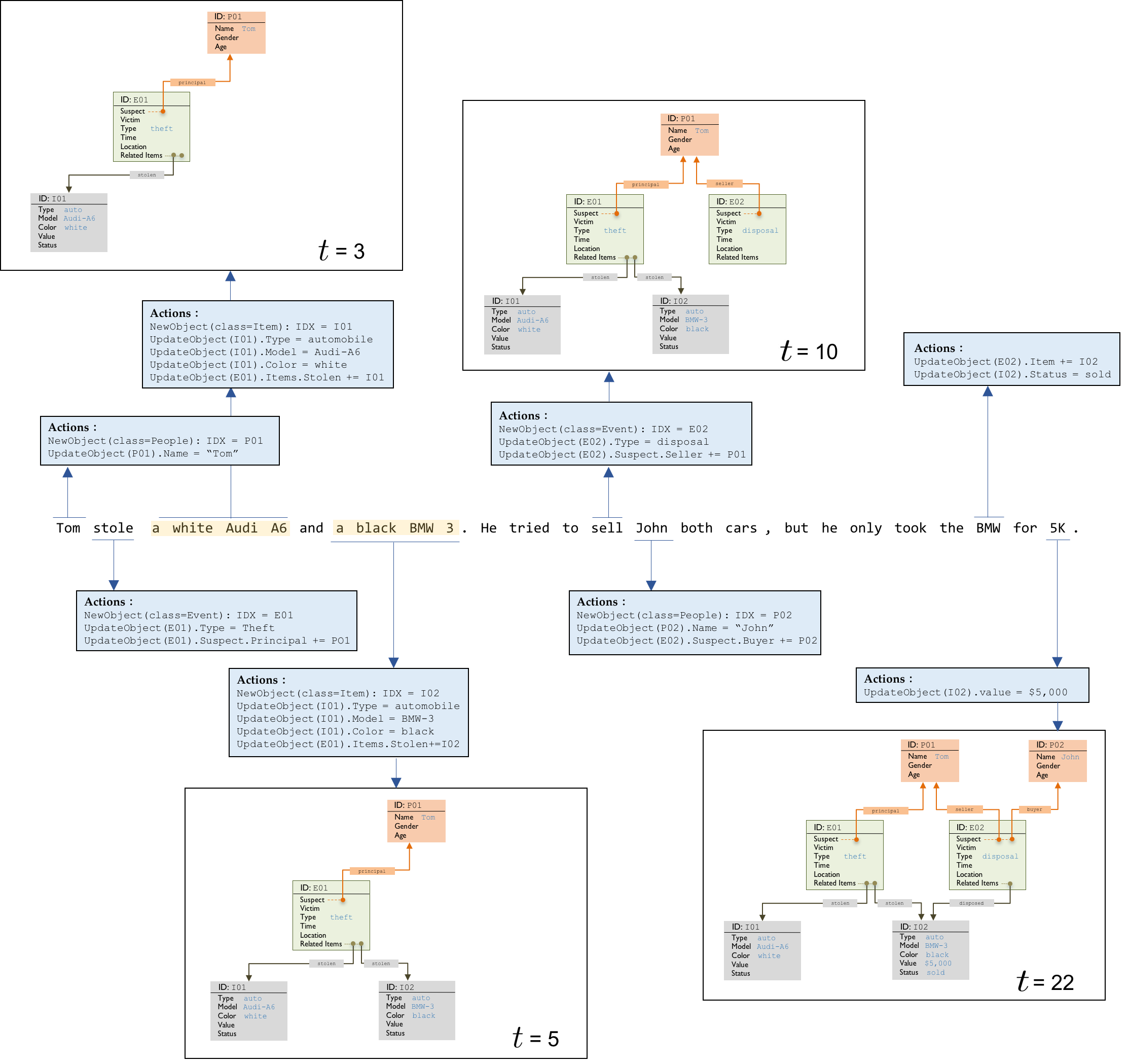}
\caption{ A pictorial illustration of a full episode of \OONP parsing, where we assume the description of cars (highlighted with shadow) are segmented in preprocessing.}
\label{f:fullEpisode}
\end{figure}

In Figure \ref{f:fullEpisode}, we give an example of the entire episode of \OONP parsing on the short text given in the example in Figure~\ref{f:OONPatWork}. Note that different from our late treatment of actions, we let some selection actions (e.g., the \assignx ) be absorbed into the updating actions to simplify the illustration.

\section{\OONPx : Neural-Symbolism}
\OONP offers a way to parse a document that imitates the cognitive process of human when reading and comprehending  a document: \OONP maintains a partial understanding of document as a mixture of symbolic  (representing clearly inferred structural knowledge) and distributed (representing knowledge without complete structure or with great uncertainty).  As shown in Figure \ref{s:components},  \Reader is taking and issuing both symbolic signals and continuous signals,  and they are entangled through \nncx.   

\OONP has plenty space for symbolic processing:  in the implementation in  Figure \ref{f:Reader2},  it is  carried out by the three symbolic processors.   For each of the symbolic processors,  the input symbolic representation could be rendered partially by neural models, therefore providing an intriguing way to entangle neural and symbolic components.  Here are three examples we implemented for two different tasks
\begin{enumerate}
\item Symbolic analysis in \textsf{Action History}:  There are many symbolic summary of history we can extracted or constructed from the sequence of actions, e.g., {\color{blue}``The system just New an object with  \textsc{Person}-class five words ago"}  or  {\color{blue} ``The system just put a paragraph starting with \texttt{`(2)'} into event-3"}.   In the implementation of \Reader shown in Figure \ref{f:Reader2},  this analysis is carried out with the component called \textsf{Symbolic Analyzer}.  Based on those more structured representation of history,  \Reader might be able to make a informed guess like {\color{blue} ``If the coming paragraph starts with \texttt{`(3)'}, we might want to put it to event-2'' } based on symbolic reasoning.  This kind of guess can be directly translated into feature to assist \Readerx 's decisions,  resembling what we do with high-order features in CRF~\cite{lafferty2001-CRF}, but the sequential decision makes it possible to construct a much richer class of features from symbolic reasoning, including those with recursive structure.  One example  of this can be found in~\cite{yanyukun17}, as a special case of \OONP on event identification. 

\item Symbolic reasoning on \omx: we can use an extra Symbolic Reasoner to take care of the high-order logic reasoning after each update of the \om caused by the actions.  This can illustrated through the following example.  \tina (a \poox) carries an \apple (an \ioox),  and \tina moves from \textsc{kitchen} (a \loox) to \garden (\loox) at time $t$.   Supposing we have both {\color{blue} \tinax-\carryx-\apple } and {\color{blue} \tinax-\locatetightx-\kitchen } relation kept in \om at time $t$,  and \OONP updates the {\color{blue} \tina-\locatetightx-\kitchen} to {\color{blue} \tina-\locatetightx-\garden} at time $t$+1, the \textsf{Symbolic Reasoner} can help to update the relation {\color{blue} \apple-\locatetightx-\kitchen } to {\color{blue} \apple-\locatetightx-\garden}. This is feasible since the \om is supposed to be logically consistent.  This external logic-based update is often necessary since it is hard to let the \nnc see the entire \om due to the difficulty to find a distributed representation of the dynamic structure there.  Please see Section \ref{s:babi} for experiments.

\item Symbolic prior in \newassign:  When \Reader determines an \newassign action,  it needs to match the information about the information at hand ($\calS_t$)  and existed objects.  There is a rich set of symbolic prior that can be added to this matching process in \sm component. For example, if $\calS_t$ contains a string labeled as entity name (in preprocessing), we can use some simple rules (part of the \sm component) to determine whether it is compatible with an object with the internal property \namex. 
\end{enumerate}

\section{Learning}
The parameters of \OONP models (denoted $\Theta$) include that for all operations and that for composing the distributed sections in \inlinex.  They can be trained with different learning paradigms:  it takes both supervised learning (SL) and reinforcement learning (RL) while allowing different ways to mix the two.  Basically,  with supervised learning,  the oracle gives the ground truth about the ``right action" at each time step during the entire decision process, with which the parameter can be tuned to maximize the likelihood of the truth. In a sense,  SL represents rather strong supervision which is related to imitation learning~\cite{Stefan99-imitation} and often requires the labeler (expert) to give not only the final truth but also when and where a decision is made.  For supervised learning,  the objective function is given as
\begin{equation}
\mathcal{J}_{\sf SL} (\Theta)= -\frac{1}{N}\sum_i^N\sum_{t=1}^{T_i} \log(\pi_t^{(i)}[a^\star_t])
\end{equation}
where $N$ stands for  the number of instances, $T_i$  stands for  the number of steps in decision process for the $i^\text{th}$ instance, $\pi_t^{(i)}[\cdot]$  stands for the probabilities of the feasible actions at $t$ from the stochastic policy, and  $a^\star_t$ stands fro the ground truth action in step $t$.

With reinforcement learning, the supervision is given as rewards during the decision process, for which an extreme case is to give the final reward at the end of the decision process  by comparing the generated ontology and the ground truth, e.g.,  \vspace{-4pt}
\begin{equation}
r_t^{(i)}=\left\{
\begin{aligned}
0,  \;\;\;&\text{if $ t\not =T_i$} \\
\textsf{match}(\textsf{M}^{T_i}_\text{obj},  \calG_i),   \;\;\;&\text{if $ t =T_i$  }\\ 
\end{aligned}
\right.
\label{e:reward} \vspace{-4pt}
\end{equation} 
where the $\textsf{match}(\textsf{M}^{T_i}_\text{obj},  \calG_i)$ measures the consistency between the ontology of in $\textsf{M}^{T_i}_\text{obj}$ and the ground truth $\calG^\star$.  We can use  any policy search algorithm to maximize the expected total reward. With the commonly used REINFORCE~\cite{williams92-reinforce} for training, the gradient is given by  \vspace{-4pt}
\begin{equation}
\nabla_\Theta\mathcal{J}_{\sf RL}(\Theta) = -\mathbb{E}_{\pi_\theta}\left[\nabla_\Theta \log \pi_\Theta \left( a_t^i | s_t^i \right) r_{t:T}^{(i)} \right] \approx -\frac{1}{NT_i}\sum_i^N\sum_{t=1}^T \nabla_\Theta \log \pi_\Theta \left( a_t^i | s_t^i \right) r_{t:T_i}^{(i)}. \vspace{-4pt}
\end{equation}

When \OONP is applied to real-world tasks, there is often quite natural SL and RL.  More specifically, for ``static ontology" one can often infer some of the right actions at certain time steps by observing the final ontology based on some basic assumption, e.g.,   \vspace{-4pt}
\begin{itemize}
\item the system should New an object the first time it is mentioned, \vspace{-4pt}
\item the system should put an extracted string (say, that for \name ) into the right property of right object at the end of the string. \vspace{-4pt}
\end{itemize}
For those that can not be fully reverse-engineered, say the categorical properties of an object (e.g., \texttt{Type} for event objects), we have to resort to RL\footnote{ 
	A more detailed exposition of this idea can be found in  \cite{liuxg2018},  where RL is used for training a multi-label classifier of text 
}  to determine the time of decision,  while we also need SL to train \pn on the content of the decision.  Fortunately it is quite straightforward to combine the two learning paradigms in optimization.  More specifically,  we maximize this combined objective  \vspace{-4pt}
\begin{equation}
\mathcal{J}(\Theta) = \mathcal{J}_{\sf SL}(\Theta)  +\lambda  \mathcal{J}_{\sf RL}(\Theta) ,  \vspace{-4pt}
\label{e:combined}
\end{equation}
where $ \mathcal{J}_{\sf SL}$ and $ \mathcal{J}_{RL}$ are over the parameters within their own supervision mode and $\lambda$ coordinates the weight of the two learning mode on the parameters they share.  Equation~\ref{e:combined}  actually indicates a deep coupling of supervised learning and reinforcement learning, since for any episode the samples of actions related to RL might affect the inputs to the models under supervised learning.


For dynamical ontology  (see Section \ref{s:babi} for example),    it is impossible to derive most of the decisions from the final ontology since they may change over time.  For those,  we have to rely mostly on the supervision at the time step to train the action (supervised mode) or count on the model to learn the dynamics of the ontology evolution by fitting the final ground truth. Both scenarios are discussed in Section \ref{s:babi} on a synthetic task.

\section{Experiments}
We applied \OONP on three document parsing tasks, to verify its efficacy on parsing documents with different characteristics and investigate different components of \OONPx.
 
\subsection{Task-I: bAbI Task}  \label{s:babi}

\subsubsection{Data and task}
\begin{figure}[h!]
	\centering
	\includegraphics[width=0.9\linewidth]{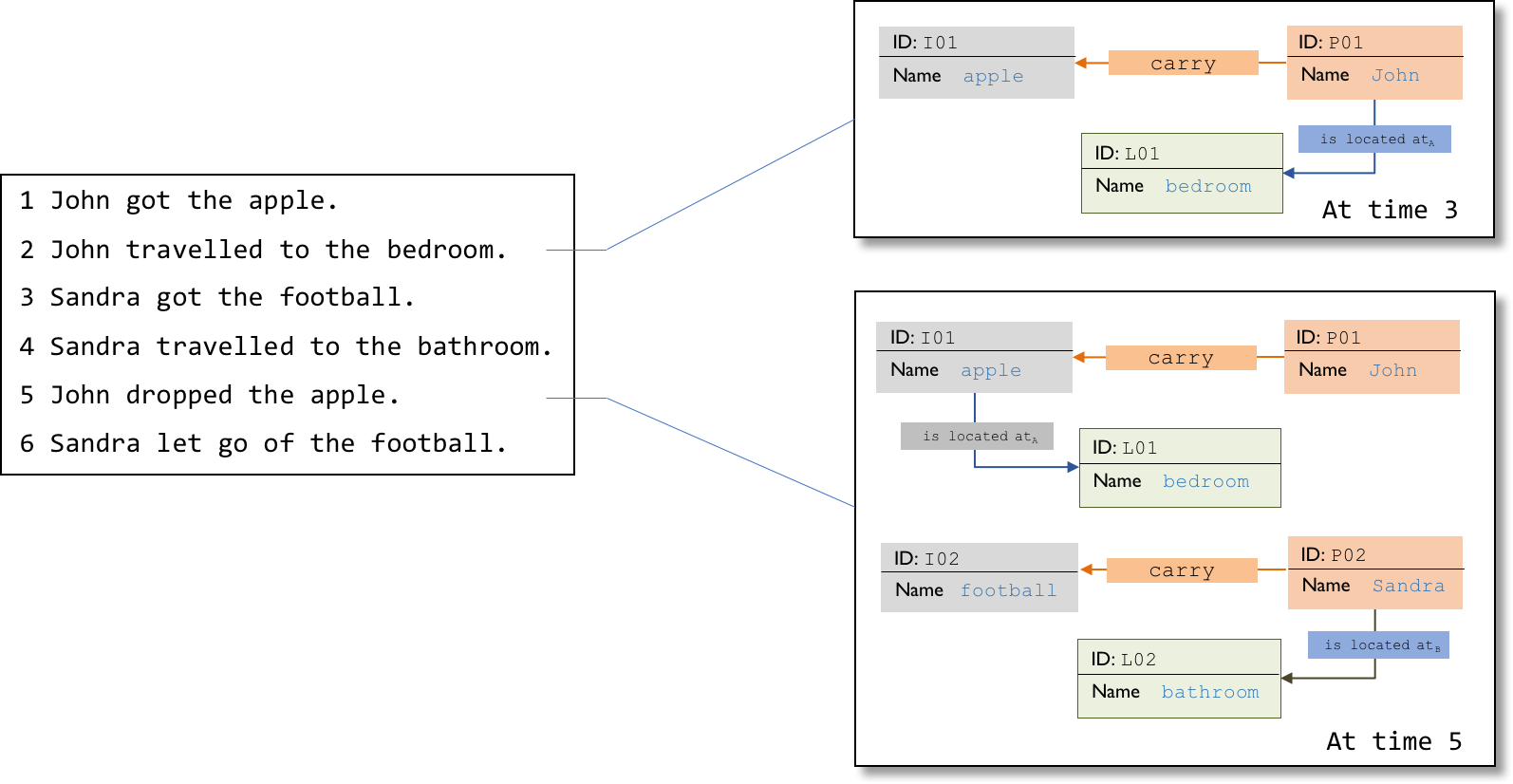}
	\vspace{-1mm}
	\caption{One instance of bAbI (6-sentence episode) and the ontology of two snapshots.}
	\label{fig:babi-example}
	\vspace{-2mm}
\end{figure}

We implemented \OONP an enriched version of bAbI tasks~\cite{Daniel-graph} with intermediate representation for history of arbitrary length. In this experiment, we considered only the original bAbi task-2~\cite{bAbItask}, with an instance shown in the left panel Figure~\ref{fig:babi-example}. 
The ontology has three types of objects: \poox, \ioox, and \loox, and three types of links:

\begin{enumerate}
\item \locatex$_A$:  between a \poo and a \loox,  \vspace{-4pt}
\item \locatex$_B$:  between a \ioo and a \loox;  \vspace{-4pt}
\item  \carryx:  between a \poo and \ioox;    \vspace{-4pt}
\end{enumerate}
which could be rendered by description of different ways.  All three types of objects have \name as the only internal property. 

The task for \OONP is to read an episode of story and recover the trajectory of the evolving ontology. We choose this synthetic dataset because it has dynamical ontology that evolves with time and  ground truth given  for each snapshot, as illustrated in Figure~\ref{fig:babi-example}. Comparing with the real-world tasks we will present later, bAbi has almost trivial internal property but relatively rich opportunities for links, considering any two objects of different types could potentially have a link.


\subsubsection{Implementation details}
For preprocessing, we have a trivial NER to find the names of people, items and locations (saved in the symbolic part of \inlinex) and word-level bi-directional GRU for the distributed representations of \inlinex. In the parsing process, \Reader goes through the inline word-by-word in the temporal order of the original text,  makes \newassign action at every word, leaving \updatewhat and \updatetowhat actions to the time steps when the read-head on \inline reaches a punctuation (see more details of actions in Table \ref{tab:actions-babi}). For this simple task, we use an almost fully neural \Reader (with MLPs for \pnx) and a vector for \matx,  with however a \sr for some logic reasoning after each update of the links, as illustrated through the following example.  Suppose at time $t$,  the ontology in \momt contains the following three facts (among others)  \vspace{-4pt}
\begin{itemize}
\item fact-1: \texttt{John} (a \poox) is in \texttt{kichten} (a \loox); \vspace{-4pt}
\item fact-2: \texttt{John}  carries \texttt{apple} (an \ioox); \vspace{-4pt}
\item fact-3: \texttt{John}  drops \texttt{apple}; \vspace{-4pt}
\end{itemize}
where fact-3 is just established by \pn at $t$.  \sr will add a new \locatex$_B$ link between \texttt{apple} and \texttt{kitchen} based on domain logic\footnote{The logic says, an item is not ``in" a location if it is held by a person.}.

\begin{table*}[h!]
	\centering
	\resizebox{0.8\linewidth}{!}{
		\begin{tabular}{|ll|}
			\hline
			\textbf{Action} & \textbf{Description}       \\
			\hline\hline
			\texttt{NewObject($c$)} & New an object of class-$c$.  \\
			\texttt{AssignObject($c,k$)}      & Assign the current information to existed object $(c,k)$ \\
			\texttt{Update($c, k$).AddLink($c',k',\ell$)}      & Add an link of type-$\ell$ from object-$(c,k)$ to object-$(c',k')$ \\
			\texttt{Update($c, k$).DelLink($c',k',\ell$)}      & Delete the link of type-$\ell$ from object-$(c,k)$ to object-$(c',k')$ \\
			\hline
		\end{tabular}
	}\vspace{-2mm}
	\caption{Actions for  bAbI.}
	\label{tab:actions-babi}
	\vspace{-2mm}
\end{table*}

\subsubsection{Results and Analysis}
For training, we use 1,000 episodes with length evenly distributed from one to six.  We use just REINFORCE with only the final reward defined as the overlap between the generated ontology and the ground truth, while step-by-step supervision on actions yields almost perfect result (result omitted). For evaluation, we use the following two metrics:
\begin{itemize}
\item the Rand index~\cite{WilliamM71-randindex} between the generated set of objects and the ground truth, which counts both the duplicate objects and missing ones, averaged over all snapshots of all test instances;
\item the F1~\cite{Rijsbergen79-F1} between the generated links and the ground truth averaged over all snapshots of all test instances, since the links are typically sparse compared with all the possible pairwise relations between objects.
\end{itemize}
with results summarized in Table~\ref{tab:acc-babi}.  \OONP can learn fairly well on recovering the evolving ontology with such a small training set and weak supervision (RL with the final reward), which clearly shows that the credit assignment over to earlier snapshots does not cause much difficulty in the learning of \OONP even with a generic policy search algorithm. It is not so surprising to observe that \sr helps to improve the results on discovering the links, while it does not improves the performance on identifying the objects although it is taken within the learning.  It is quite interesting to observe that \OONP achieves rather high accuracy on discovering the links while it performs relatively poorly on specifying the objects. It is probably due to the fact that the rewards does not penalizes the objects.

\begin{table*}[!hbtp]
 \centering
 \resizebox{0.65\linewidth}{!}{
  \begin{tabular}{|l|c|c|c|c|c|c|c|c|}
   \hline
   {\small model} &  {\small F1 (for links) \%}  & {\small RandIndex (for objects)\%}\\
   \hline
   \hline
\OONP ({\small without \textsf{S.R.}})      & 94.80 & 87.48 \\
   \hline
 	\OONP ({\small with \textsf{S.R.} }) & 95.30 & 87.48  \\
   \hline
  \end{tabular}
 }\vspace{-2mm}
 \caption{The performance a implementation of \OONP on bAbI task 2.}
 \label{tab:acc-babi}
 \vspace{-2mm}
\end{table*}

\subsection{Task-II: Parsing Police Report} \label{s:e2}
\subsubsection{Data \&  task}
\begin{figure}[h!]
	\centering
	\includegraphics[width=0.85\linewidth]{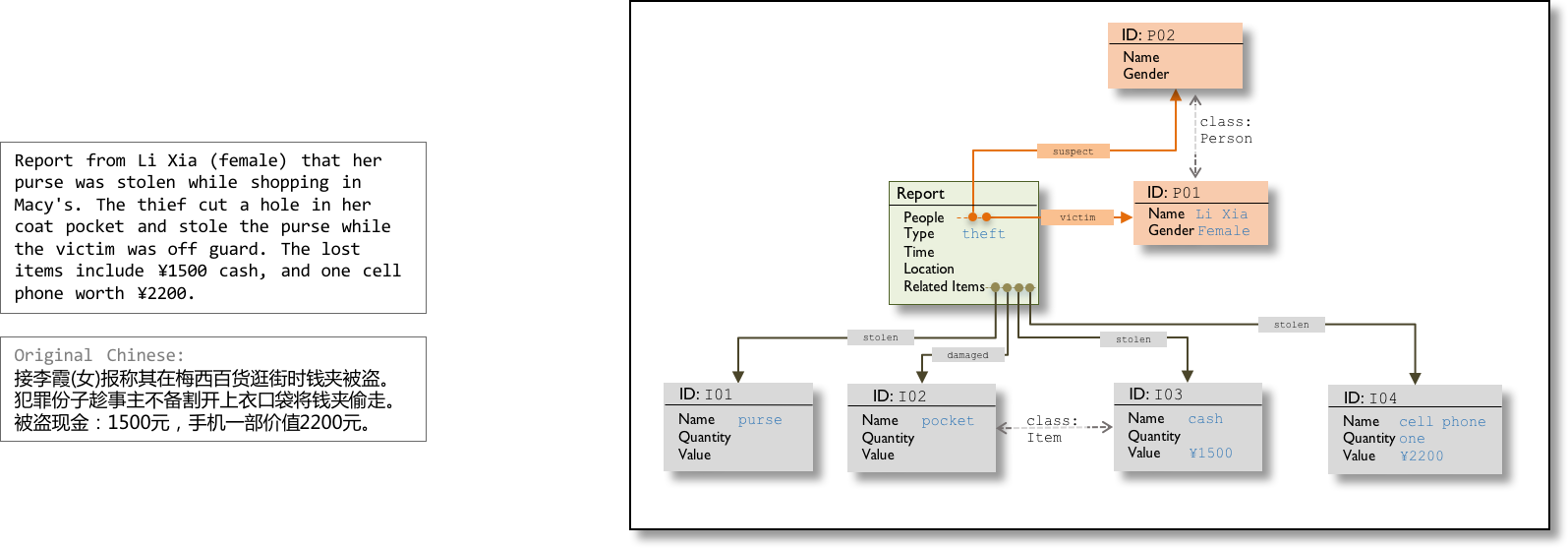}
	\caption{An example of police report and its ontology.}\label{fig:task-example}
\end{figure}

We implement \OONP for parsing  Chinese police report (brief description of criminal cases written by policeman), as illustrated in the left panel of Figure  \ref{fig:task-example}.   We consider a corpus of 5,500 cases with a variety of crime categories,  including theft, robbery, drug dealing and others.  The ontology we designed for this task mainly consists of a number of  \poos and \ioos connected through a \eoo with several types of relations, as illustrated in the right panel of Figure  \ref{fig:task-example}.  A \poo has three internal properties: \name (string),  \gender (categorical) and \age (number), and two types of external links (\suspect and \victimx) to an \eoox.  An \ioo has three internal properties: \name (string), \quantity (string) and \valuee (string), and six types of external links  (\texttt{stolen}, \texttt{drug}, \texttt{robbed}, \texttt{swindled}, \texttt{damaged}, and \texttt{other}) to an \eoox.   Compared with bAbI in Section \ref{s:babi}, the police report ontology has less pairwise links but much richer internal properties for objects of all three objects.  Although the language in this dataset is reasonably formal,  the corpus coverages a big variety of topics and language styles, and has a high proportion of typos. On average, a sample has 95.24 Chinese words and the ontology has 3.35 objects,  3.47 mentions and 5.02 relationships. The average length of a document is 95 Chinese characters, with digit string (say, ID  number)  counted as one character.

\subsubsection{Implementation details}

 The \OONP model is designed to generate ontology as illustrated in Figure  \ref{fig:task-example}  through a decision process with actions in  Table \ref{tab:actions-Police2}.   As pre-processing, we performed regular NER with third party algorithm (therefore not part of the learning) and simple rule-based extraction to yield the symbolic part of \inline as shown in Figure \ref{fig:inline_sym}.  For the distributed part of \inlinex, we used dilated CNN with different choices of depth and kernel size~\cite{Yu15-dilated-conv},  all of which will be jointly learned during training.  In making the \newassign decision,  \Reader considers the matching between  two structured objects, as well as the hints from the symbolic part of \inline as features, as pictorially illustrated in Figure \ref{f:compare}.  In updating objects with its string-type properties (e.g., \name for a \poo ), we use Copy-Paste strategy for extracted string (whose NER tag already specifies which property in  an object  it goes to) as \Reader sees it.  For undetermined category properties in existed objects,  \pn will determine the object to update (an \newassign action without \neww option), its property to update (an \updatewhat action), and the updating operation (an \updatetowhat action) at milestones of the decision process , e.g., when reaching an punctuation. For this task, since all the relations are between the single by-default \eoo and other objects,  the relations can be reduced to category-type properties of the corresponding objects in practice.  For category-type properties,  we cannot recover  \newassign   and \updatewhat actions from the label (the final ontology), so we resort RL for learning to determine that part, which is mixed with the supervised learning for \updatetowhat and other actions for string-type properties.  

\begin{table*}[h!]
	\centering
	\resizebox{1\linewidth}{!}{
		\begin{tabular}{|ll|}
			\hline
			\textbf{Action} & \textbf{Description}       \\
			\hline\hline
			\texttt{NewObject($c$)} & New an object of class-$c$.  \\
			\texttt{AssignObject($c,k$)}      & Assign the current information to existed object $(c,k)$ \\
			\texttt{UpdateObject($c,k$).Name}      & Set the name of object-$(c,k)$ with the extracted string.\\
			\texttt{UpdateObject($\pppx,k$).Gender}      & Set the name of a \poo indexed $k$ with the extracted string.\\
			\texttt{UpdateObject($\iiix,k$).Quantity}  & Set the quantity of an \ioo indexed $k$  with the extracted string.\\
			\texttt{UpdateObject($\iiix,k$).Value}  & Set the value of an  \ioo indexed $k$ with the extracted string.\\
			\texttt{UpdateObject($\eeex,1$).Items.x} & Set the link between the \eoo and an \ioox, where  
			 \texttt{x} $\in$\{\texttt{stolen}, \\& \texttt{drug}, \texttt{robbed}, \texttt{swindled}, \texttt{damaged}, \texttt{other}\}\\
			\texttt{UpdateObject($\eeex,1$).Persons.x} & Set the link between the \eoo and an \poox, 
			                                                        and \texttt{x} $\in$\{\victimx, \\& \suspectx \}\\
			\hline
		\end{tabular}
	}\vspace{-2mm}
	\caption{Actions for parsing police report.}\label{tab:actions-Police2}
	\vspace{-2mm}
\end{table*}

\subsubsection{Results \& discussion} \label{s:e2-sect3}
We use 4,250 cases for training, 750 for validation an held-out 750 for test.  We consider the following four metrics in comparing the performance of different models: 
\vspace{-5pt}
\begin{table*}[h!]
 \centering
 \resizebox{0.95\linewidth}{!}{
  \begin{tabular}{ll}
   Assignment  Accuracy  & the accuracy on \newassign actions made by the model \\ 
 Category Accuracy & the accuracy of predicting the category properties of all the objects\\ 
 Ontology Accuracy   & the proportion of instances for which the generated ontology is  exactly  \\& the same as the ground truth \\   
  Ontology Accuracy-95    & the proportion of instances for which the generated ontology achieves \\&  95\% consistency with the ground truth \\      
  \end{tabular}
 }\vspace{-2mm}
 \vspace{-2mm}
\end{table*}

\noindent which measures the accuracy of the model in making discrete decisions as well as generating the final ontology.  We empirically examined several \OONP implementations and compared them with two baselines, Bi-LSTM and EntityNLM \cite{ji2017entitynlm}, with results given in { Table~\ref{tab:resultsx}}.

\begin{table}[h!]
	\centering
	\resizebox{0.9\linewidth}{!}{
		\begin{tabular}{|l|c|c|c|c|}
			\hline
			Model &   Assign Acc. (\%)  & Type Acc. (\%)  &   Ont. Acc. (\%)  & Ont. Acc-95 (\%) \\
			\hline
			\hline
			Bi-LSTM (baseline) &  73.2 $\pm$ 0.58 &- & 36.4$\pm$ 1.56 & 59.8 $\pm$ 0.83\\
			\hline
			ENTITYNLM (baseline) & 87.6 $\pm$ 0.50 & 84.3 $\pm$ 0.80  & 59.6 $\pm$ 0.85  & 72.3  $\pm$ 1.37 \\
			\hline \hline
			\OONP (neural)  & 88.5 $\pm$ 0.44 & 84.3 $\pm$ 0.58  & 61.4 $\pm$ 1.26 & 75.2 $\pm$ 1.35 \\
			\hline
			\OONP (structured) &  91.2 $\pm$ 0.62 & 87.0 $\pm$ 0.40 & 65.4 $\pm$ 1.42 & 79.9 $\pm$ 1.28 \\
			\hline
			\OONP (RL)  & \textbf{91.4} $\pm$ 0.38 & \textbf{87.8} $\pm$ 0.75 & \textbf{66.7} $\pm$ 0.95 & \textbf{80.7} $\pm$ 0.82  \\
			\hline
		\end{tabular} }
		\caption{\OONP on parsing police reports.  
		} 
		\label{tab:resultsx}
		\vspace{-2mm}
	\end{table}

The Bi-LSTM and EntityNLM are essentially a simple version of \OONP without a structured \carryon and designed operations (sophisticated matching function in  \newassign). Basically the Bi-LSTM baseline consists of a Bi-LSTM \inline encoder and a two-layer MLP on top of that acting as a simple \pn for prediction actions.  Since this baseline does not has an explicit object representation,  it does not support category type of prediction. We hence only train this baseline model to perform \newassign actions, and evaluate with the Assignment Accuracy (first metric)  and a modified version of Ontology Accuracy (third and fourth metric) that counts only the properties that can be predicted, hence in favor of Bi-LSTM. As for EntityNLM, as another strong baseline,  it can model an arbitrary number of entities in context while generating each entity mention at an arbitrary length and perform well in coreference resolution, and entity prediction~\cite{ji2017entitynlm}. Adapted to this scenario, it is re-implemented to predict object index and properties of object with a minor change that name prediction task is replaced by the identical third-party algorithm for fairness. We consider three \OONP variants:

\begin{itemize}
\item \OONP(neural): simple version of \OONP with only distributed representation in \Reader in determining all actions;
\item \OONP(structured): \OONP that considers the matching between two structured objects in \newassign actions, with symbolic prior encoded in \sm and other features for \pnx;
\item \OONP(RL):  another version of \OONP(structured) that uses RL to determine the time for predicting the category properties, while  \OONP(neural) and \OONP(neural) use a rule-based approach to determine the time.
\end{itemize}

As shown in Table~\ref{tab:resultsx}, Bi-LSTM baseline struggles to achieve around 73\% Assignment Accuracy on test set, while \OONP(neural) can boost the performance to 88.5\%. Arguably, this difference in performance is due to the fact that Bi-LSTM lacks \omx, so all relevant information has to be stored in the Bi-LSTM hidden states along the reading process. When we start putting symbolic representation and operation into \Reader, as shown in the result of \OONP (structure),  the performance is again significantly improved on all four metrics. More specifically, we have the following two observations (not shown in the table), 
\begin{itemize}
\item Adding inline symbolic features as in Figure \ref{fig:inline_sym} improves around 0.5\% in \newassign action prediction, and 2\% in category property prediction. The features we use include the type of the candidate strings and the relative distance to the maker character we chose.
\begin{figure}[h!]
	\centering
	\includegraphics[width=0.5\linewidth]{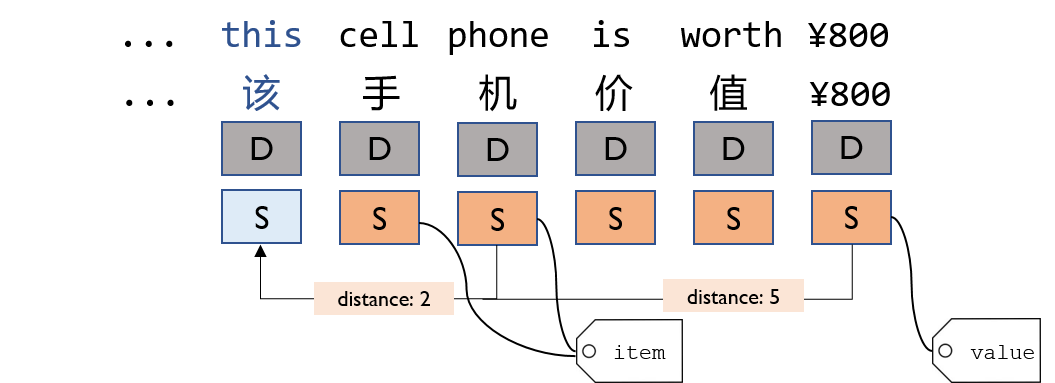}
	\vspace{-1mm}
	\caption{Information in distributed and symbolic forms in \inlinex.}\label{fig:inline_sym}
	\vspace{-2mm}
\end{figure}
\item Using a matching function that can take advantage of the structures in objects helps better generalization. Since the objects in this task has multiple property slots like \texttt{Name, Gender, Quantity, Value}. We tried adding both the original text (e.g., \texttt{Name, Gender, Quantity, Value} ) string of an property slot and the embedding of that as additional features, e.g.,  the length the longest common string between the candidate string and a relevant property of  the object.
\end{itemize}

When using REINFORCE to determine when to make prediction for category property, as shown in the result of \OONP (RL),  the prediction accuracy  for  category property and the overall ontology accuracy is improved. It is quite interesting that it  has some positive impact on the supervised learning task (i.e., learning the \newassign actions)  through shared parameters. The entanglement of the two learning paradigms in \OONP is one topic for future research, e.g., the effect of predicting the right category property on the \newassign actions if the predicted category property is among the features  of the matching function for \newassign actions.

%

\subsection{Task-III: Parsing court judgment documents} \label{s:e3}
\subsubsection{Data and task}
\begin{figure}[h!]
	\centering
	\includegraphics[width=1\linewidth]{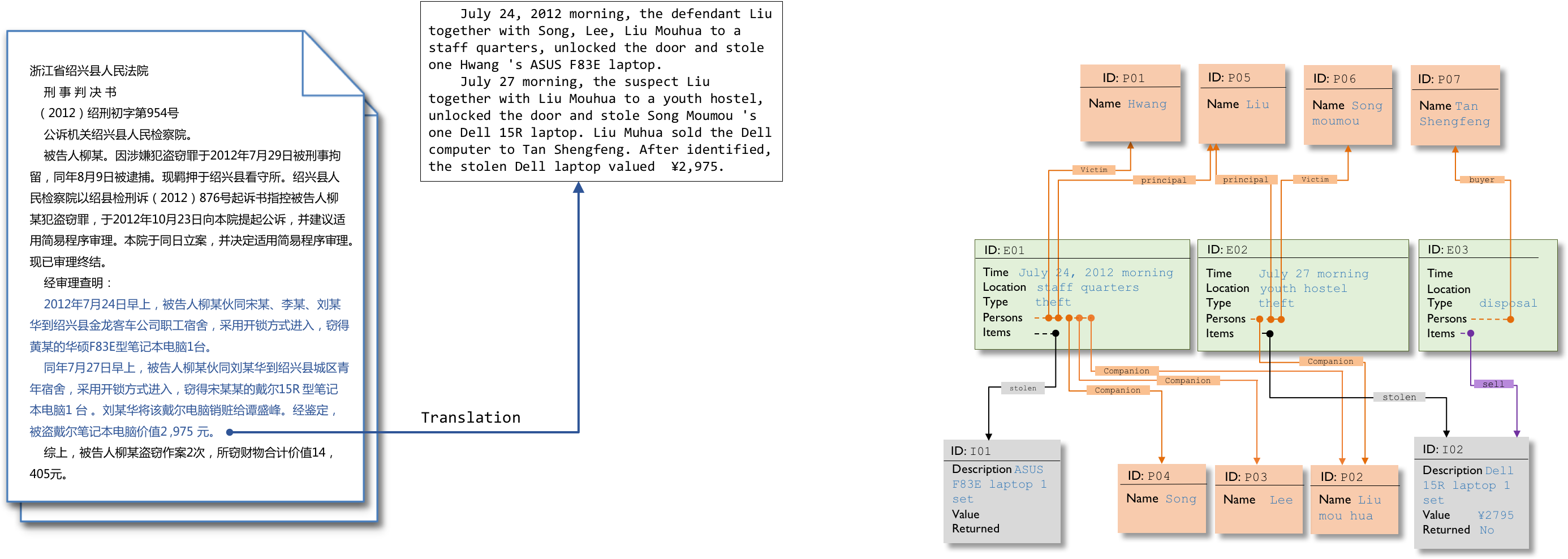}
	\vspace{-18pt}
	\caption{Left panel: the judgement document with highlighted part being the description the facts of crime, right panel: the corresponding ontology}
	\label{fig:judgement-example}
	\vspace{-2mm}
\end{figure}

We also implement \OONP for parsing  court judgement on theft. Unlike the two previous tasks, court judgements are typically much longer, containing multiple events of different types as well as bulks of irrelevant text, as illustrated in the left panel of Figure  \ref{fig:task-example}.   The dataset contains 1961 Chinese judgement documents, divided into training/dev/testing set with 1561/200/200 texts respectively. The ontology we designed for this task mainly consists of a number of  \poos and \ioos connected through a number \eoo with several types of links.  A \eoo has three internal properties: \timee (string), \location (string), and \typee (category, $\in$\{\theftx, \restix, \dispx \}), four types of external links to \poos (namely, \texttt{principal}, \texttt{companion}, \texttt{buyer}, \victimx)  and four types of external links to \ioos (\texttt{stolen},  \texttt{damaged}, \texttt{restituted}, \texttt{disposed}).  In addition to the external links to \eoos, a \poo has only the \name (string) as the internal property.  An \ioo has three internal properties: \desc  (array of strings), \valuee (string) and \returned (binary) in addition to its external links to \eoos, where \desc consists of the words describing the corresponding item, which could come from multiple segments across the document.  A \poo or an \ioo could be linked to more than one \eoox, for example a person could be the principal suspect in event $A$ and also a companion in event $B$. An illustration of the judgement document and the corresponding ontology can be found in Figure~\ref{fig:judgement-example}. 
%
%
%

 \subsubsection{Implementation details }

 We use a model configuration similar to that in Section \ref{s:e2}, with however the following important difference.  In this experiment, \OONP performs a 2-round reading on the text. In the first round, \OONP identifies the relevant events,  creates empty \eoox s, and does Notes-Taking on \inline to save the information about event segmentation (see~\cite{yanyukun17} for more details). In the second round, \OONP read the updated \inlinex,  fills the \eoox s, creates and fills \poox s and \ioox s, and specifies the links between them.  When an object is created during a certain event,  it will be given an extra feature (not an internal propoerty) indicating this connection, which will be used in deciding links between this object and event object, as well as in determining the future \newassign actions.  The actions of the two round reading are summarized in Table~\ref{tab:actions-Judge}.

\begin{table*}[h!]
	\centering
	\resizebox{1\linewidth}{!}{
		\begin{tabular}{|ll|}
			\hline
			\textbf{Action for 1st-round} & \textbf{Description}       \\
			\hline
		       \texttt{NewObject($c$)} & New an \eoox, with $c=$\eeex.  \\
		       \texttt{NotesTaking($\eeex, k$).word} & Put indicator of event-$k$ on the current word.  \\
		       \texttt{NotesTaking($\eeex, k$).sentence} & Put indicator of event-$k$ on the rest of sentence, and move the read-head to the  \\
		        & first word of next sentence  \\
		       \texttt{NotesTaking($\eeex, k$).paragraph} & Put indicator of event-$k$ on the rest of the paragraph, and move the read-head to \\
		       &  the first word of next paragraph.  \\
                      \texttt{Skip.word} & Move the read-head  to next word \\
		       \texttt{Skip.sentence} & Move the read-head  to the first word of next sentence  \\
		       \texttt{Skip.paragraph} & Move the read-head  to the first word of next paragraph.  \\
		       \hline\hline
		       			\textbf{Action for 2nd-round} & \textbf{Description}       \\
		       \hline 
			\texttt{NewObject($c$)} & New an object of class-$c$.  \\
			\texttt{AssignObject($c,k$)}      & Assign the current information to existed object $(c,k)$ \\
			\texttt{UpdateObject($\pppx,k$).Name}      & Set the name of the $k^\text{th}$ \poo with the extracted string.\\
			\texttt{UpdateObject($\iiix,k$).Description}  & Add to the description of an $k^\text{th}$  \ioo with the extracted string.\\
			\texttt{UpdateObject($\iiix,k$).Value}  & Set the value of an $k^\text{th}$  \ioo with the extracted string.\\
			\texttt{UpdateObject($\eeex,k$).Time}  & Set the time of an $k^\text{th}$  \eoo with the extracted string.\\
		       \texttt{UpdateObject($\eeex,k$).Location}  & Set the location of an $k^\text{th}$  \eoo with the extracted string.\\
			\texttt{UpdateObject($\eeex,k$).Type} & Set the type of the $k^\text{th}$ \eoo  among  
			  \{\texttt{theft},     \texttt{disposal},  \texttt{restitution}\}\\
			\texttt{UpdateObject($\eeex,k$).Items.x} & Set the link between the $k^\text{th}$ \eoo  and an \ioox, where  
			 \texttt{x} $\in$ \\&  \{\texttt{stolen},  \texttt{damaged}, \texttt{restituted}, \texttt{disposed} \}\\
			 \texttt{UpdateObject($\eeex,k$).Persons.x} & Set the link between the  $k^\text{th}$ \eoo and an \poox, 
			                                                        and \texttt{x} $\in$ \\&\{\principalx, \companionx, \buyerx, \victimx \}\\
			\hline
		\end{tabular}
	}\vspace{-2mm}
	\caption{Actions for parsing court judgements.}\label{tab:actions-Judge}
	\vspace{-2mm}
\end{table*}


\subsubsection{Results and Analysis}
We use the same metric as in  Section \ref{s:e2}, and compare two \OONP variants, \OONP(neural) and \OONP(structured),  with two baselines, EntityNLM and Bi-LSTM. The two baselines will be tested only on the second-round reading, while both \OONP variant are tested on a two-round  reading.  
The results are shown in Table~\ref{tab:judge-acc}. \OONP parsers attain  accuracy significantly higher than Bi-LSTM models. Among, \OONP (structure) achieves over 64\% accuracy on getting the entire ontology right and over 78\% accuracy on getting 95\% consistency with the ground truth.

\begin{table}[h!]
	\centering 
	\resizebox{0.9\linewidth}{!}{
		\begin{tabular}{|l|c|c|c|c|}
			\hline
			Model &   Assign Acc. (\%)  & Type Acc. (\%)  &   Ont. Acc. (\%)  & Ont. Acc-95 (\%) \\
			\hline
			\hline
			Bi-LSTM (baseline) & 84.66  $\pm$ 0.20 & - & 18.20 $\pm$ 0.74 & 36.88 $\pm$ 1.01 \\
			\hline
			ENTITYNLM (baseline) & 90.50 $\pm$ 0.21   & 96.33 $\pm$ 0.39 & 39.85 $\pm$ 0.20 & 48.29 $\pm$ 1.96\\
			\hline
			\OONP(neural) & 94.50 $\pm$ 0.24  & 97.73 $\pm$ 0.12 & 53.29 $\pm$ 0.26 & 72.22 $\pm$ 1.01\\
			\hline
			\OONP(structured) & \textbf{96.90} $\pm$ 0.22 & \textbf{98.80} $\pm$ 0.08  & \textbf{71.11}  $\pm$ 0.54 & \textbf{77.27} $\pm$ 1.05\\
			\hline
		\end{tabular}}  
		\caption{\OONP on judgement documents.}
		\label{tab:judge-acc}
		\vspace{-5mm}
	\end{table}
\section{Conclusion}
We proposed Object-oriented Neural Programming (\OONPx),  a framework for semantically parsing in-domain documents. \OONP  is neural net-based, but equipped with sophisticated architecture and mechanism for document understanding, therefore nicely combining interpretability and learnability.  Experiments on both synthetic and real-world datasets have shown that \OONP outperforms several strong baselines by a large margin on parsing fairly complicated ontology.

\section*{Acknowledgments}
We thank Fandong Meng and Hao Xiong for their insightful discussion. We also thank Classic Law Institute for providing the raw data.

\bibliographystyle{apalike}
\bibliography{cite}
\end{document}